\documentclass[letterpaper, 10 pt, conference]{ieeeconf}  

\IEEEoverridecommandlockouts                              

\usepackage{amsmath,amssymb,amsfonts}

\DeclareMathOperator*{\argmin}{argmin}

\usepackage{siunitx} 

\usepackage{algorithmicx}
\usepackage{algorithm}
\usepackage{algpseudocode}

\makeatletter
\renewcommand{\ALG@beginalgorithmic}{\small}
\makeatother

\usepackage{graphicx}
\usepackage{subcaption}
\usepackage{textcomp}

\usepackage{booktabs}       
\usepackage{multirow}       
\usepackage{multicol}       
\usepackage{gensymb}

\usepackage{xcolor}
\def\BibTeX{{\rm B\kern-.05em{\sc i\kern-.025em b}\kern-.08em
    T\kern-.1667em\lower.7ex\hbox{E}\kern-.125emX}}

\allowdisplaybreaks

\newcommand\scalemath[2]{\scalebox{#1}{\mbox{\ensuremath{\displaystyle #2}}}}

\title{
\LARGE \bf Robust, High-Precision GNSS Carrier-Phase Positioning\\ with Visual-Inertial Fusion}

\author{Erqun Dong$^{1,2*}$, Sheroze Sheriffdeen$^{2}$, Shichao Yang$^{2}$, Jing Dong$^{2}$, Renzo De Nardi$^{2}$, Carl Ren$^{2}$,\\Xiao-Wen Chang$^{1}$, Xue Liu$^{1}$ and Zijian Wang$^{2}$
\thanks{*This work was performed during an internship at Meta.}
\thanks{$^{1}$School of Computer Science, McGill University, Montreal, Quebec, Canada,
        {\tt\small erqun.dong@mail.mcgill.ca, \{chang, xueliu\}@cs.mcgill.ca}}%
\thanks{$^{2}$Meta, Reality Labs Research, Redmond, WA, USA
        {\tt\small \{sheroze, shichaoy, jingdong, renzo, carlren, zjwang\}@fb.com}}%
}

\begin{document}

\maketitle
\thispagestyle{empty}
\pagestyle{empty}

\begin{abstract}

Robust, high-precision global localization is fundamental to a wide range of outdoor robotics applications. Conventional fusion methods use low-accuracy pseudorange based GNSS measurements ($>5m$ errors) and can only yield a coarse registration to the global earth-centered-earth-fixed (ECEF) frame. In this paper, we leverage high-precision GNSS carrier-phase positioning and aid it with local visual-inertial odometry (VIO) tracking using an extended Kalman filter (EKF) framework that better resolves the integer ambiguity concerned with GNSS carrier-phase. 
We also propose an algorithm for accurate GNSS-antenna-to-IMU extrinsics calibration to accurately align VIO to the ECEF frame. Together, our system achieves robust global positioning demonstrated by real-world hardware experiments in severely occluded urban canyons, and outperforms the state-of-the-art RTKLIB by a significant margin in terms of integer ambiguity solution fix rate and positioning RMSE accuracy. 
\end{abstract}


\section{Introduction}
In this paper, we tackle the robustness issue of GNSS carrier-phase-based global positioning in challenging urban canyons with the aid of VIO.

Robust, centimeter-level global localization is a cornerstone in many outdoor applications such as mobile robots, self-driving cars, and augmented reality.
Currently, a large family of localization solutions rely on visual-inertial odometry (VIO)~\cite{mourikis2007multi, qin2018vins, campos2021orb} because of the accurate short-distance tracking.
Without pre-built maps that are costly to build and maintain, VIO has limited usage in large-scale applications that require accurate global positioning. 
Inspired by some recent works that fuse GNSS pseudorange positioning (GNSS-PR) into VIO to mitigate VIO's drift~\cite{cioffi2020tightly, lee2020intermittent, qin2019general, cao2022gvins, boche2022visual}, we fuse VIO into GNSS carrier-phase positioning (GNSS-CP) towards a low-cost but high-precision global positioning for outdoor applications.
\begin{figure}
\centering
\begin{subfigure}{0.9\linewidth}
\includegraphics[width=\linewidth]{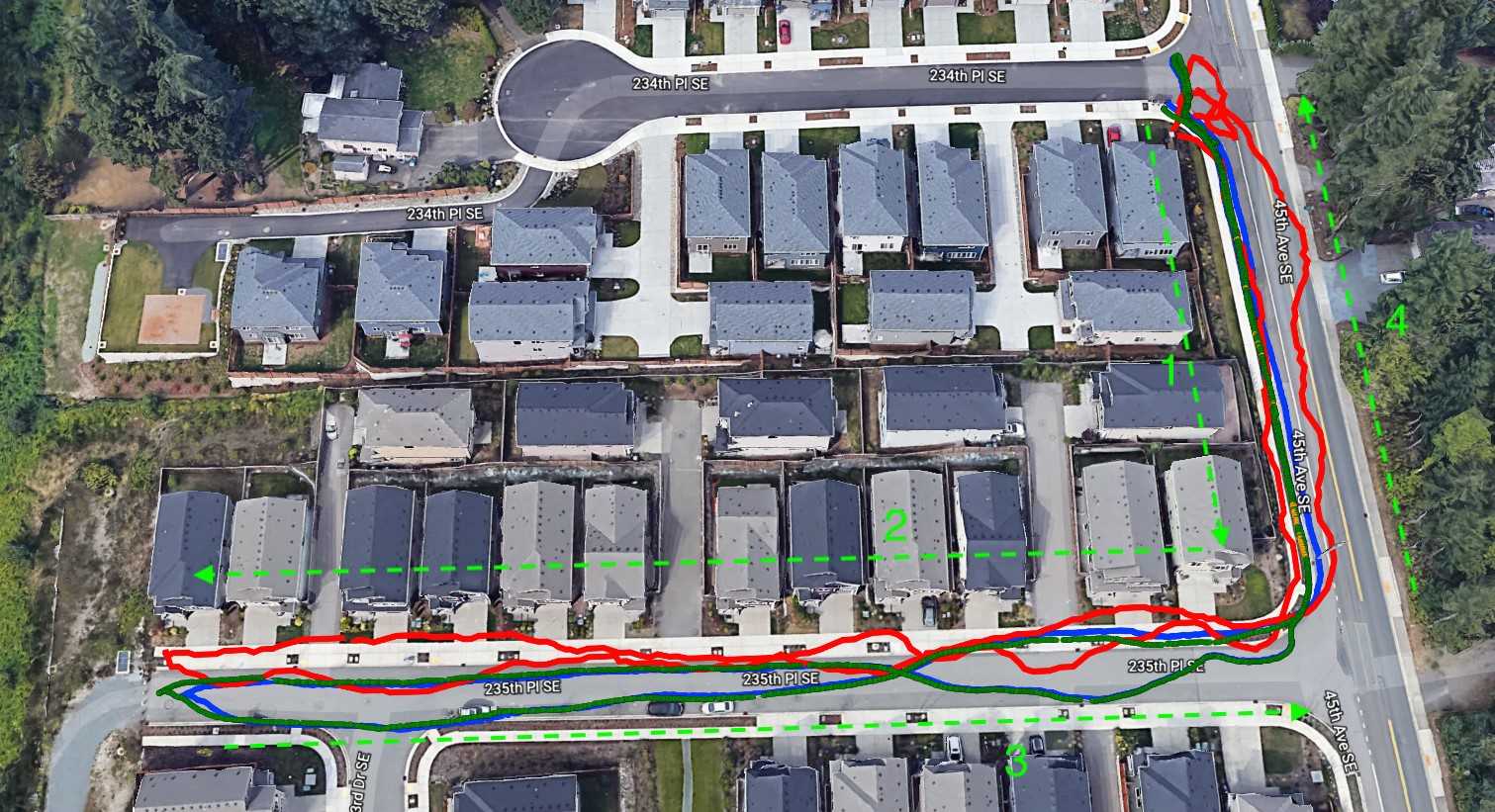}
\end{subfigure}
\caption{Demonstration of GNSS pseudorange trajectory (red), GNSS carrier-phase trajectory (green) and VIO trajectory aligned with GNSS carrier-phase in the first \SI{50}{\meter} (blue). The light-green numbers and dashed lines indicate the direction of four major parts of the trajectory. We can observe that the pseudorange trajectory is much noisier than the other two with a positioning error of up to \SI{10}{\meter}.
%
This indicates the potential of GNSS carrier-phase for more precise global positioning than GNSS pseudorange.}
\label{fig:qualitative}
\end{figure}

GNSS-CP has manifested ability for drift-free centimeter-level localization of reported 2--5cm accuracy under open sky.
In contrast, GNSS-PR with $>$5m accuracy has limited potential for centimeter-level localization. See Fig.~\ref{fig:qualitative} for a demonstration. %
%
%
However, GNSS-CP is more susceptible and fragile to line-of-sight blockage and multi-path disturbance in dense urban areas.
Section V-D shows that the accuracy of GNSS-CP in urban canyons can degrade from centimeter-level to be even tens of meters.
So towards robust GNSS carrier-phase-based global positioning in challenging urban canyons, we address these points in this paper: 
1) leveraging the local tracking and uncertainty estimation from VIO to better condition and constrain the search space of integer least sqaures, and
2) from an engineering perspective, reject measurements from outlier satellites in urban environments so that the GNSS carrier-phase resolution are less prone to noisy multi-path signals.
%

%
To achieve this, we carefully register the VIO trajectories to the ECEF frame with a proposed calibration algorithm to solve the translation offset between the IMU and the center of the GNSS antenna.
We loosely couple VIO and GNSS-CP in an EKF framework, where VIO serves as the motion model that propagates prior state estimates and covariances, and GNSS-CP serves as the measurement update to obtain a posterior float solution.
We design strict outlier rejection to tackle signal blockage and multi-path reflection in challenging environments such as urban canyons.
Altogether, the integer ambiguity resolution can be seeded with good float solutions (where the integer ambiguities are treated as real numbers) to solve for fixed solutions (where the integer ambiguities are fixed as integers) with a higher success rate.

We conduct hardware experiments in urban canyons using the Aria glasses \cite{aria} with a rigidly attached, low-cost, multi-band GNSS antenna.
Results show that our calibration renders centimeter-level alignment error, and the overall positioning and integer ambiguity fixing rate of our method outperforms the state-of-the-art GNSS-CP named RTKLIB~\cite{takasu2009development}.
It is worth mentioning that the urban canyons in our experiments (Fig.~\ref{fig:downtown},~\ref{fig:library}) are severer than that in the experiments of recent papers~\cite{li2019tight, liu2021optimization, yoder2020multi, cao2022gvins} in terms of the heights and density of the buildings. %

\section{Related Work}
The fusion of visual-inertial information with GNSS has been explored since the success of visual and visual-inertial localization and mapping~\cite{mourikis2007multi,raul2016orbslam2,qin2018vins}.
Some works~\cite{yu2019gps, qin2019general, cioffi2020tightly, lee2020intermittent, liu2021optimization, boche2022visual} localize in the local ENU coordinate system by a fusion of visual-inertial with GNSS pseudorange positioning.
Others can align the local odometry coordinate frame onto the global ECEF coordinate frame with GNSS pseudorange measurements~\cite{mascaro2018gomsf, cao2022gvins}.
They model either pseudorange positioning results or pseudorange measurements as Gaussian noises.
However, due to the noisiness of GNSS pseudorange measurements/positioning, the alignment of VIO and GNSS are subject to large errors in the range of several meters even under the open sky (refer to Table II in~\cite{cao2022gvins}). 
Importantly, most work that fuses pseudorange GNSS measurement and visual inertial sensing (such as~\cite{lee2020intermittent,qin2019general, boche2022visual}) report results against vanilla RTK or RTKLIB as groundtruth, whereas in this paper we achieve better performance than the RTK groundtruth as used by others.

%
%
There were prior works that attempted global positioning with GNSS carrier-phase, camera, and IMU.
%
Li et al. \cite{li2019tight} and Yoder et al. \cite{yoder2020multi} fused visual-inertial information as a prior estimate to aid the integer ambiguity resolution. 
Shepard et al. \cite{shepard2014high} used a bundle adjustment method for estimation.
These papers assumed that the IMU-Antenna extrinsics calibration parameters are known constants beforehand without characterizing their values.
In contrast, we propose a lightweight optimization method to estimate it and experimentally validate it.
%
%
Moreover, all three works lack real-world experiments in severely challenging urban canyons like ours.
%
Moreover, \cite{yoder2020multi} has the limitation of requiring two GNSS antennas installed on the device.

Research efforts have been made to assist carrier phase positioning in unfavorable sky visibility condition.
Takasu et al.~\cite{takasu2008cycle} use dead-reckoning by IMU to take over the navigation.
Building heights constructed by Lidar~\cite{wen2018correcting} are used to reduce satellite signal outliers.
Thresholding-based outlier detection methods are also used to mitigate challenging urban canyon effects~\cite{li2019tight, liu2021optimization, yoder2020multi, cao2022gvins}, but the real-world environments in their experiments are not as challenging as that in ours.
In our paper, we consider VIO's incremental motion as a strong prior and reject outliers in both measurement update and integer ambiguity resolution. %

\section{Problem Formulation}
%
%

Suppose there are $g$ groups of signals received by our GNSS module and the reference station. 
Each group are in the same constellation and have the same frequency, and suppose group $j$ has $m_j$ satellites.
%
Then the GNSS measurement at a certain epoch is $\boldsymbol y = [\boldsymbol {\phi}_{1}^\top, \dots, \boldsymbol {\phi}_{g}^\top, \boldsymbol {\rho}_{1}^\top, \dots, \boldsymbol {\rho}_g^\top]^\top$, where $\boldsymbol {\phi}_{j}$ is the double-differenced carrier-phase measurements \cite{misra2006global} in the $j$-th group, and $\boldsymbol {\rho}_j$ is the double-differenced pseudorange measurements in the $j$-th group.
With one satellite selected as the reference satellite for double-differencing, $\boldsymbol \phi_j$ and $\boldsymbol \rho_j$ both have $(m_j - 1)$ dimensions.

The visual-inertial inputs are processed by VIO providing IMU-centric device poses $\{\mathrm I\}$ in the local odometry frame $\{\mathrm O\}$, denoted as $\boldsymbol T _{\scriptscriptstyle \mathrm{OI}}\in \mathrm{SE}(3)$ where $\mathrm{SE}(3)$ is the three-dimensional Special Euclidean Group. The VIO also outputs associated uncertainty~\cite{mourikis2007multi}.
Then we interpolate the VIO trajectory at the timestamps of GNSS carrier-phase measurements for time-interpolated VIO poses.
We take a loosely-coupled fusion approach by feeding the output of VIO unilaterally to our GNSS-CP algorithm. 
%
%
%
%
%
%
%
We assume the intrinsics and extrinsics of cameras and IMUs are assumed to be calibrated, and the extrinsics between GNSS antenna and IMU, denoted $\boldsymbol t _{\scriptscriptstyle \mathrm{IA}}\in \mathbb{R}^3$, need to be estimated.
%

%
%
%
We assume that an initial convergence of carrier-phase positioning (not necessarily fixed solution) is available before our fusion algorithm starts, which allows us to initially align the VIO local odometry frame to the ECEF frame (shown in Fig.~\ref{fig:downtown-our} and Fig.~\ref{fig:library-our}) to warm start our algorithm.
%
%

%
The state to estimate is denoted
$\boldsymbol x=[{}^{\scriptscriptstyle \mathrm E} {\boldsymbol p}_{\scriptscriptstyle \mathrm{u}}^\top,\boldsymbol d_{1}^\top, \dots, \boldsymbol d_{g}^\top]^\top$,
where ${}^{\scriptscriptstyle \mathrm E} {\boldsymbol p}_{\scriptscriptstyle \mathrm{u}}$ means the current position of the user in the ECEF frame $\{\mathrm E\}$,
and 
%
%
%
$\boldsymbol d_{j} = [d_{j}^{(1,2)}, d_{j}^{(1,3)},\dots,d_{j}^{(1,m_j)}]^\top$ is the double-differenced integer ambiguities for satellites in group $j$. 
%
%
%
%
%
We build our fusion system upon a baseline EKF named RTKLIB~\cite{takasu2009development}.
The prediction step of the baseline EKF uses a constant velocity model; whereas we use VIO for better kinematic modeling, introduced in Section~\ref{subsec:fusion}.
For the measurement update step, the measurement equations are double-differenced carrier-phase and pseudorange measurements.
Specifically, for each group $j$,
\begin{align*}
 \boldsymbol \phi_{j} = \boldsymbol h_{\phi,j}(\boldsymbol x) + \boldsymbol \epsilon_{\phi,j}, \ \ \ \
 \boldsymbol \rho_{j} = \boldsymbol h_{\rho,j}(\boldsymbol x) + \boldsymbol \epsilon_{\rho,j},
\end{align*}
%
where $\boldsymbol \epsilon_{\phi,j}$ is the Gaussian noise of the double-differenced carrier-phase measurements and $\boldsymbol \epsilon_{\rho,j}$ is the Gaussian noise of the double-differenced pseudorange measurements (For the formulation of $\boldsymbol \epsilon_{\phi,j}$ and $\boldsymbol \epsilon_{\rho,j}$, readers are referred to~\cite{misra2006global}, Section 7.5).
%
The indivisual measurement functions in $\boldsymbol h_{\phi,j}(\boldsymbol x)$ and $\boldsymbol h_{\rho,j}(\boldsymbol x)$ are 
%
%
%
$$\scalemath{0.83}{
    h_{\phi,j}^{(1, s)}(\boldsymbol x) = (||{}^{\scriptscriptstyle \mathrm E} {\boldsymbol t}_{\scriptscriptstyle \mathrm{u}}^{(1)}||_2 - ||{}^{\scriptscriptstyle \mathrm E} {\boldsymbol t}_{\scriptscriptstyle \mathrm{r}}^{(1)}||_2) - (||{}^{\scriptscriptstyle \mathrm E} {\boldsymbol t}_{\scriptscriptstyle \mathrm{u}}^{(s)}||_2 - ||{}^{\scriptscriptstyle \mathrm E} {\boldsymbol t}_{\scriptscriptstyle \mathrm{r}}^{(s)}||_2)
    + \lambda_j d_{j}^{(1,s)}} ,
$$
$$\scalemath{0.9}{
    h_{\rho,j}^{(1, s)}(\boldsymbol x) = (||{}^{\scriptscriptstyle \mathrm E} {\boldsymbol t}_{\scriptscriptstyle \mathrm{u}}^{(1)}||_2 - ||{}^{\scriptscriptstyle \mathrm E} {\boldsymbol t}_{\scriptscriptstyle \mathrm{r}}^{(1)}||_2) - (||{}^{\scriptscriptstyle \mathrm E} {\boldsymbol t}_{\scriptscriptstyle \mathrm{u}}^{(s)}||_2 - ||{}^{\scriptscriptstyle \mathrm E} {\boldsymbol t}_{\scriptscriptstyle \mathrm{r}}^{(s)}||_2)} ,
 $$
where subscripts $\mathrm u$ and $\mathrm r$ mean the user and the reference station, superscripts $1$ and $s$ are satellite ID's where without loss of generality we assume the reference satellite's ID is $1$,
$\lambda_j$ means the wavelength of group $j$'s signal, $||{}^{\scriptscriptstyle \mathrm E} {\boldsymbol t}_{\scriptscriptstyle \mathrm{u}}^{(1)}||_2$ means the distance between the user $\mathrm u$ and satellite $1$ satisfying $||{}^{\scriptscriptstyle \mathrm E} {\boldsymbol t}_{\scriptscriptstyle \mathrm{u}}^{(1)}||_2 = || {}^{\scriptscriptstyle \mathrm E} {\boldsymbol p}_{\scriptscriptstyle \mathrm{u}} - {}^{\scriptscriptstyle \mathrm E} {\boldsymbol p}^{(1)} ||_2$, and similar meanings apply to $||{}^{\scriptscriptstyle \mathrm E} {\boldsymbol t}_{\scriptscriptstyle \mathrm{r}}^{(1)}||_2$, $||{}^{\scriptscriptstyle \mathrm E} {\boldsymbol t}_{\scriptscriptstyle \mathrm{u}}^{(s)}||_2$, and $||{}^{\scriptscriptstyle \mathrm E} {\boldsymbol t}_{\scriptscriptstyle \mathrm{r}}^{(s)}||_2$. 


%
The EKF measurement update treats all components of $\boldsymbol x$, including both position variables and integer ambiguities, as real numbers to solve a float solution~\cite{misra2006global}.
%
Based on the float solution, we then use the LAMBDA algorithm~\cite{lambda,mlambda} to resolve the fixed solution, where a core step is solving (searching) an integer least squares problem
\begin{align}
    \boldsymbol D = \argmin_{\boldsymbol D \in \mathbb Z^N}\  (\boldsymbol D - \hat {\boldsymbol D})^\top \boldsymbol W_{\hat {\boldsymbol D}}^{-1} (\boldsymbol D - \hat {\boldsymbol D}),
\label{eq:ils}
\end{align}
where $\boldsymbol D = [\boldsymbol d_{1}^\top, \dots, \boldsymbol d_{g}^\top]^\top$ is all the double-differenced integer ambiguities in all groups, $\hat {\boldsymbol D}$ is its float solution, $N$ is its dimension, and $\boldsymbol W_{\hat {\boldsymbol D}}^{-1}$ is the inverse of the covariance of $\hat {\boldsymbol D}$.
After integer ambiguity is searched and validated with the ratio test~\cite{verhagen2013ratio}, we can get a fixed solution of $^{\scriptscriptstyle \mathrm E} {\boldsymbol p}_{\scriptscriptstyle \mathrm{u}}$ that is more accurate than the float solution~\cite{lambda}.
In this integer least-squares problem, the float solution and its covariance form an ellipsoidal search space for the correct integer, so their quality affects the success rate of the integer ambiguity search, and equivalently affects the fixed solution rate.

In a nutshell, this paper calculates the state $\boldsymbol x$ at every epoch with the GNSS measurements $\boldsymbol y$ at each epoch and the VIO poses $\boldsymbol T_{OI}$.

\section{Method}

%
%
We achieve the robust positioning by improving the quality of the float solution and covariance in Eq. (\ref{eq:ils}) for better integer ambiguity resolution with a fusion of VIO into GNSS carrier-phase.
Well-aligned VIO can be much more accurate than RTKLIB's constant velocity model in terms of both prior prediction and covariance propagation. 
Plus, the accuracy of VIO tracking can also effectively reject outlier satellites so that only the line-of-sight GNSS signals are used for positioning.
We introduce these methods in this section, and demonstrate the system performance in the experiments Section~\ref{subsec:simulation} and ~\ref{subsec:uc}.
%
%

%
\subsection{Alignment of VIO and GNSS with Extrinsics Calibration}
\label{subsec:calib}
We obtain the transformation from the odometry frame $\{\mathrm O\}$ to the ECEF frame $\{\mathrm E\}$, $\boldsymbol T _{\scriptscriptstyle \mathrm{EO}}$, by aligning the VIO trajectory with the GNSS trajectory with a least squares optimization. 
For precise alignment, the extrinsic parameter $\boldsymbol t _{\scriptscriptstyle \mathrm{IA}}$, which represents the translation between the center of the IMU and the center of the GNSS antenna, should also be considered.
Given a GNSS trajectory consisting of a sequence of the user's 3D positions ${}^{\scriptscriptstyle \mathrm E} {\boldsymbol p}_{\scriptscriptstyle \mathrm{u}_i},\ i=1,\dots,k$, and the corresponding time-interpolated VIO trajectory ${\boldsymbol T} _{\scriptscriptstyle \mathrm{OI}_i},\ i=1,\dots,k$, 
%
%
%
we optimize the transformation $\boldsymbol T _{\scriptscriptstyle \mathrm{EO}}$ and the extrinsic parameter $\boldsymbol t _{\scriptscriptstyle \mathrm{IA}}$ with
\begin{align}
\label{eq:calib}
    \min_{\substack{\boldsymbol T _{\scriptscriptstyle \mathrm{EO}}\in \mathrm{SE}(3),\  \\ \boldsymbol t _{\scriptscriptstyle \mathrm{IA}}\in \mathbb{R}^3 }} \sum_{i=1}^{k} \left\lVert \boldsymbol T _{\scriptscriptstyle \mathrm{EO}} \cdot {\boldsymbol T} _{\scriptscriptstyle \mathrm{OI}_i} \cdot \begin{bmatrix} 
      \boldsymbol t _{\scriptscriptstyle \mathrm{IA}}\\
      1
    \end{bmatrix} - {}^{\scriptscriptstyle \mathrm E} {\boldsymbol p}_{\scriptscriptstyle \mathrm{u}_i} \right\rVert  ^2.
\end{align}
%
%
For the alignment ${\boldsymbol T} _{\scriptscriptstyle \mathrm{EO}}$ to be well constrained on all degrees of freedom, the trajectories needs persistence of excitation~\cite{persistenceOfExcitation} along translation and rotation components.
To this end, we use principal component analysis \cite{hotelling1933analysis} to decide if the first eigenvalue ratio (FER) to the sum of all eigenvalues is smaller than a threshold. 


%

%
We solve the optimization Eq. (\ref{eq:calib}) using a two-pass process.
First, we initialize $\boldsymbol T _{\scriptscriptstyle \mathrm{EO}}$ to identity, fix $\boldsymbol t _{\scriptscriptstyle \mathrm{IA}}$ to 0, and optimize Eq. \eqref{eq:calib} for a coarse estimate of $\boldsymbol T _{\scriptscriptstyle \mathrm{EO}}$.
%
%
Second, we initialize 
$\boldsymbol T _{\scriptscriptstyle \mathrm{EO}}$ to the coarse estimate from the first pass, and optimize Eq. \eqref{eq:calib} again for $\boldsymbol T _{\scriptscriptstyle \mathrm{EO}}$ and $\boldsymbol t _{\scriptscriptstyle \mathrm{IA}}$ jointly.
%
The two-pass process is necessary for ensuring that the nonlinear optimization is seeded with a good initialization to avoid bad local minimum.
%
%
%
%
Section~\ref{subsec:calib-on-alignment} demonstrates the improvement of alignment accuracy with our proposed calibration method.
%
%

%
Considering VIO's drifting nature, we need to frequently re-align VIO with the ECEF frame.
All the later alignments optimize ${\boldsymbol T} _{\scriptscriptstyle \mathrm{EO}}$ in Eq.~\ref{eq:calib} with $\boldsymbol t _{\scriptscriptstyle \mathrm{IA}}$ fixed.
In real-world urban canyon environments, GNSS positioning can degrade severely and impede accurate trajectory alignment.
Therefore, we use an root mean square error (RMSE) threshold of 0.1m and an FER threshold of 97\% to decide whether to perform a new trajectory alignment at a new epoch.
The frequency of re-alignment is annotated in Fig.~\ref{fig:downtown-our} and Fig.~\ref{fig:library-our}.

\subsection{Fusion Algorithm}
\label{subsec:fusion}
With the VIO registered to the ECEF frame, we then fuse VIO with GNSS in a loosely-coupled fashion.
We use VIO to track the incremental motion in ECEF as the EKF's prior prediction and propagates covariance.
Then we follow the routine in RTKLIB~\cite{takasu2009development} to resolve integer ambiguity with the MLAMBDA algorithm~\cite{mlambda}.
Major designs of our system are introduced in this section.
The pseudocode for the overall fusion routine is summarized in Algorithm~\ref{alg:fusion}.

{\centering
\resizebox{0.87\linewidth}{!}{
\begin{minipage}{\linewidth}
\begin{algorithm}[H]
\caption{Fusion Algorithm for Each Epoch}
\label{alg:fusion}
\begin{algorithmic}[1]
\State NewEpochData in epoch $i+1$ captured
\State $\hat {\boldsymbol T} _{\scriptscriptstyle \mathrm{EO}} \gets$AlignNewSegmentationOrReuseOldAlignment()
\State ${}^{\scriptscriptstyle \mathrm O} \boldsymbol t_{\scriptscriptstyle \mathrm {I}_{i} \scriptscriptstyle \mathrm {I}_{i+1}}\gets$ VIO(NewEpochData) 
\State ${}^{\scriptscriptstyle \mathrm E} {\boldsymbol p}_{\scriptscriptstyle \mathrm{u}_{i+1|i}}\gets$Eq. (\ref{eq:state-prior})
\State $\boldsymbol P_{i+1|i}\gets$Eq. (\ref{eq:cov-prior})
\State $\bar {\boldsymbol y}\gets$OutlierSatellitePruning(NewEpochData)
\State Float solution$\gets$MeasurementUpdate($\bar{\boldsymbol y},\boldsymbol x$)
\If {Float solution validation passed}
\State Fixed solution$\gets$ MLAMBDA($\bar {\boldsymbol y}$)
\If {Fixed solution validation passed}
\State Final solution$\gets$Fixed solution and covariance
\Else 
\State $\bar {\boldsymbol y}'\gets$Further prune satellites that produce large residuals in integer least squares
\State Fixed solution$\gets$ MLAMBDA($\bar {\boldsymbol y}'$)
\If {Fixed solution validation passed}
\State Final solution$\gets$Fixed solution and covariance
\Else
\State Final solution$\gets$Float solution and covariance
\EndIf
\EndIf
\Else
\State Final solution$\gets$the prior ${}^{\scriptscriptstyle \mathrm E} {\boldsymbol p}_{\scriptscriptstyle \mathrm{u}_{i+1|i}},\boldsymbol P_{i+1|i}$
\EndIf
\State Return Final solution
\end{algorithmic}
\end{algorithm}
\end{minipage}
}
\par
}

\smallskip
\subsubsection{EKF Prior Prediction}
\label{sec:vio-pred-ekf}
%
%
%
Suppose the state at epoch $i$ is already estimated and we are to estimate the state at epoch $i+1$.
Since VIO has a higher frequency than GNSS, we seek in the VIO's sliding window for two poses with the nearest capture times to GNSS epochs $i$ and $i+1$, denoted $\boldsymbol T _{\scriptscriptstyle \mathrm{OI}_{i}}$ and $\boldsymbol T _{\scriptscriptstyle \mathrm{OI}_{i+1}}$.
Then the prior position estimate, ${}^{\scriptscriptstyle \mathrm E} {\boldsymbol p}_{\scriptscriptstyle \mathrm{u}_{i+1|i}}$, can be predicted by the position estimate of the previous epoch, ${}^{\scriptscriptstyle \mathrm E} {\boldsymbol p}_{\scriptscriptstyle \mathrm{u}_{i|i}}$, and the incremental translation tracked by VIO in the odometry frame, ${}^{\scriptscriptstyle \mathrm O} \boldsymbol t_{\scriptscriptstyle \mathrm {I}_{i} \scriptscriptstyle \mathrm {I}_{i+1}}$, with
\begin{align}
 \label{eq:state-prior}
     {}^{\scriptscriptstyle \mathrm E} {\boldsymbol p}_{\scriptscriptstyle \mathrm{u}_{i+1|i}} &=  {}^{\scriptscriptstyle \mathrm E} {\boldsymbol p}_{\scriptscriptstyle \mathrm{u}_{i|i}} + \boldsymbol R _{\scriptscriptstyle \mathrm{EO}} \cdot {}^{\scriptscriptstyle \mathrm O} \boldsymbol t_{\scriptscriptstyle \mathrm {I}_{i} \scriptscriptstyle \mathrm {I}_{i+1}},
\end{align}
where $\boldsymbol R _{\scriptscriptstyle \mathrm{EO}}$ is the rotation part of $\boldsymbol T _{\scriptscriptstyle \mathrm{EO}}$.
%
%
%
%

%
\subsubsection{Covariance Propagation}
Denote the prior covariance of ${}^{\scriptscriptstyle \mathrm E} {\boldsymbol p}_{\scriptscriptstyle \mathrm{u}_{i+1}}$ as $\boldsymbol P_{i+1|i}$, and the posterior covariance of ${}^{\scriptscriptstyle \mathrm E} {\boldsymbol p}_{\scriptscriptstyle \mathrm{u}_{i}}$ as $\boldsymbol P_{i|i}$.
We propagate $\boldsymbol P_{i+1|i}$ from  $\boldsymbol P_{i|i}$ combining VIO's covariance.
We obtain the $12\times 12$ covariance matrix 
$\mathop{\mathrm{cov}}(\begin{bmatrix}\boldsymbol \xi _{\scriptscriptstyle \mathrm{OI}_{i}}\\ \boldsymbol \xi _{\scriptscriptstyle \mathrm{OI}_{i+1}}\end{bmatrix})$
from the covariance of the joint Gaussian distribution maintained by VIO, where $\boldsymbol \xi _{\scriptscriptstyle \mathrm{OI}_{i}}$ and $\boldsymbol \xi _{\scriptscriptstyle \mathrm{OI}_{i+1}}$ are the corresponding Lie algebra elements~\cite{sola2018micro} for $\boldsymbol T _{\scriptscriptstyle \mathrm{OI}_{i}}$ and $\boldsymbol T _{\scriptscriptstyle \mathrm{OI}_{i+1}}$.
%
%
Then the covariance of the incremental translation in the odometry frame is
%
\begin{align}
\label{eq:relative-cov}
    \mathop{\mathrm{cov}}({}^{\scriptscriptstyle \mathrm O} \boldsymbol t_{\scriptscriptstyle \mathrm {I}_{i} \scriptscriptstyle \mathrm {I}_{i+1}}) =  [\boldsymbol J_{\scriptscriptstyle \mathrm{I}_{i}} \boldsymbol J_{\scriptscriptstyle \mathrm{I}_{i+1}}] \mathop{\mathrm{cov}}(\begin{bmatrix}\boldsymbol \xi _{\scriptscriptstyle \mathrm{OI}_{i}}\\ \boldsymbol \xi _{\scriptscriptstyle \mathrm{OI}_{i+1}}\end{bmatrix})  \begin{bmatrix} 
      \boldsymbol J_{\scriptscriptstyle \mathrm{I}_{i}}^\top\\
      \boldsymbol J_{\scriptscriptstyle \mathrm{I}_{i+1}}^\top
    \end{bmatrix},
\end{align}
where $\boldsymbol J_{\scriptscriptstyle \mathrm{I}_{i}}$ and $\boldsymbol J_{\scriptscriptstyle \mathrm{I}_{i+1}}$ are the partial derivatives of ${}^{\scriptscriptstyle \mathrm O} \boldsymbol t_{\scriptscriptstyle \mathrm {I}_{i} \scriptscriptstyle \mathrm {I}_{i+1}}$ w.r.t. the Lie algebra element of $\boldsymbol T _{\scriptscriptstyle \mathrm{OI}_{i}}$ and $\boldsymbol T _{\scriptscriptstyle \mathrm{OI}_{i+1}}$, respectively.
In Eq. (\ref{eq:relative-cov}), $\mathop{\mathrm{cov}}({}^{\scriptscriptstyle \mathrm O} \boldsymbol t_{\scriptscriptstyle \mathrm {I}_{i} \scriptscriptstyle \mathrm {I}_{i+1}})$ is $3\times 3$, and $\boldsymbol J_{\scriptscriptstyle \mathrm{I}_{i}}$ and $\boldsymbol J_{\scriptscriptstyle \mathrm{I}_{i+1}}$ are both $3\times 6$.
%
%
Then in the ECEF frame,
\begin{align}
\mathrm{cov}({}^{\scriptscriptstyle \mathrm E} \boldsymbol t_{\scriptscriptstyle \mathrm {I}_{i} \scriptscriptstyle \mathrm {I}_{i+1}}) =  \boldsymbol R _{\scriptscriptstyle \mathrm{EO}} \cdot \mathrm{cov}({}^{\scriptscriptstyle \mathrm O} \boldsymbol t_{\scriptscriptstyle \mathrm {I}_{i} \scriptscriptstyle \mathrm {I}_{i+1}}) \cdot \boldsymbol R _{\scriptscriptstyle \mathrm{EO}}^\top.
\end{align}
The final form of the proposed covariance propagation of the GNSS prior position in EKF prediction step is
\begin{align}
    \boldsymbol P_{i+1|i} = \boldsymbol P_{i|i} + \mathrm{cov}({}^{\scriptscriptstyle \mathrm E} \boldsymbol t_{\scriptscriptstyle \mathrm {I}_{i} \scriptscriptstyle \mathrm {I}_{i+1}}). \label{eq:cov-prior}
\end{align}
\subsubsection{Outlier Satellite Pruning}
\label{subsubsec:outlier}
%
%
In practice, as GNSS signals undergo severe blockage and reflection in urban canyons, signals traveling along non-line-of-sight (NLoS) paths should not participate in the positioning.
Therefore, we need to detect and rule out such outlier satellite measurements.
We do outlier rejection in both the measurement update step and the integer ambiguity resolution step.
In the measurement update step, since VIO provides accurate EKF prediction, we use 5 times of VIO's incremental motion as a threshold to filter the innovation of the satellite measurements.
The remaining measurements $\bar {\boldsymbol y}$ is a subset $\boldsymbol y$.
%
%
%
%
After this, some elements of $\bar {\boldsymbol y}$ can still be unreliable with some NLoS signals passing the threshold.
%
%
%
%
%
%
%
Then in the integer ambiguity resolution step, if the integer ambiguities solved by the MLAMBDA~\cite{mlambda} algorithm cannot pass the validation by the ratio test~\cite{verhagen2013ratio}, the system further prunes satellites that produce largest residuals in the integer least squares problem, and then solve it again.
This way, some more spurious outliers can be ruled out.
\begin{figure}[!tbp]
  \centering
  \subfloat[Aria glasses with external dual-band GNSS antenna.]{\includegraphics[height=0.34\linewidth]{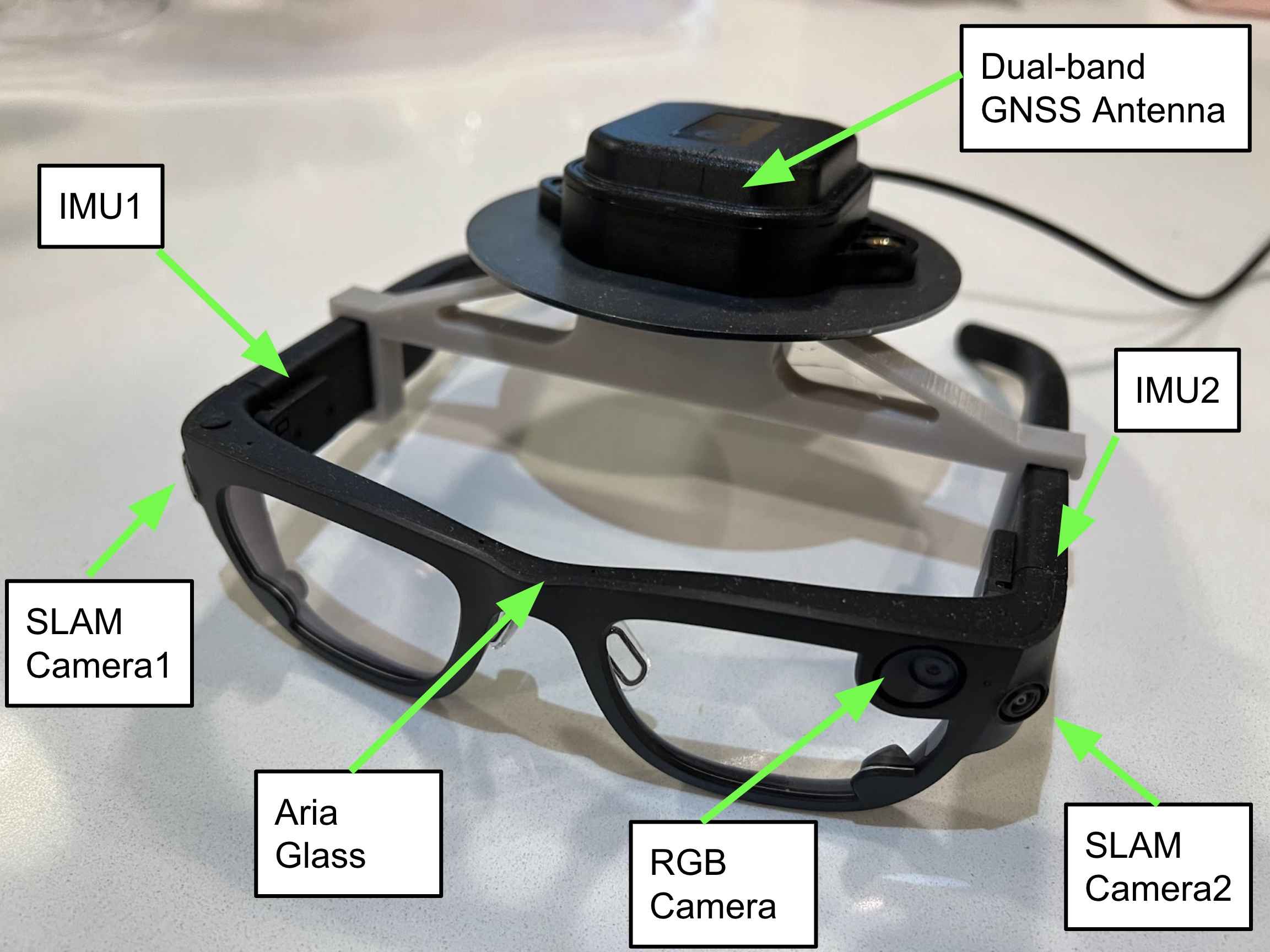}\label{fig:hardware-aria}}
  \hfill
  \subfloat[GNSS receiver board.]{\includegraphics[height=0.34\linewidth]{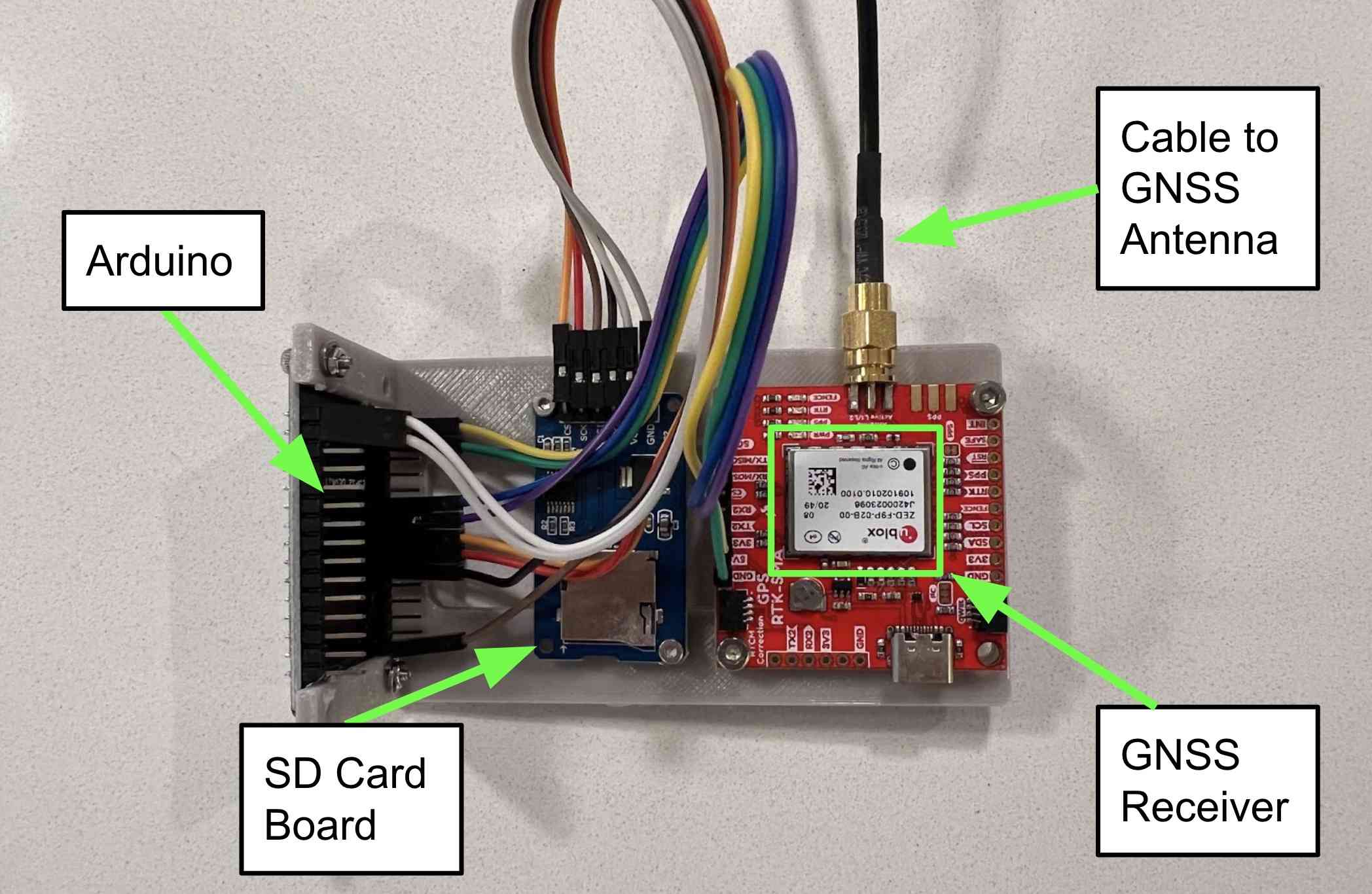}\label{fig:hardware-arduino}}
  \caption{Device used in the hardware experiments.}
\end{figure}

\section{Experiments}
\subsection{Hardware Implementation}

We perform hardware experiments in real-world challenging urban canyons (Bellevue, Washington, USA) using the Aria glasses~\cite{aria}, which has a suite of machine perception sensors including two global-shutter monochrome SLAM cameras, two 6-DOF IMUs, and a built-in GPS pseudorange module.
%
%
%
We rigidly attach an external multi-band GNSS carrier-phase antenna to the Aria glasses with a metal ground plate, as shown in Fig.~\ref{fig:hardware-aria}.
The GNSS carrier-phase measurements are received by a u-BLox ZED-F9P GNSS receiver, and the raw data is further logged by an Arduino into an SD-card for offline processing\footnote{Our algorithm is casual, which is the same as EKF.}.
We use three constellations in our experiments: GPS, GLONASS, and Galileo.
We set up our own reference station using the same antenna and receiver, about \SI{20}{\km} from the user's receiver.
This baseline length is common in many cities in north America between a public base station and its nearby urban environment.
The two SLAM cameras operate at \SI{10}{\hertz} at $640\times480$ resolution, and the two IMUs operate at \SI{1000}{\hertz} and \SI{800}{\hertz}, respectively. The internal GPS pseudorange module runs at \SI{1}{\hertz} while the external GNSS carrier-phase module at \SI{5}{\hertz}.
For synchronization between GNSS carrier-phase and VIO, since the built-in GPS of the Aria glasses and the external GNSS carrier-phase antenna both record GPS Time, we can simply use the built-in GPS of Aria glasses as a bridge to synchronize the VIO (system time on Aria glasses) and the external GNSS carrier-phase signals.

\begin{figure*}[t!]
 \centering
  \begin{subfigure}{0.33\linewidth}
     \centering
     \includegraphics[width=\linewidth,height=0.7\linewidth]{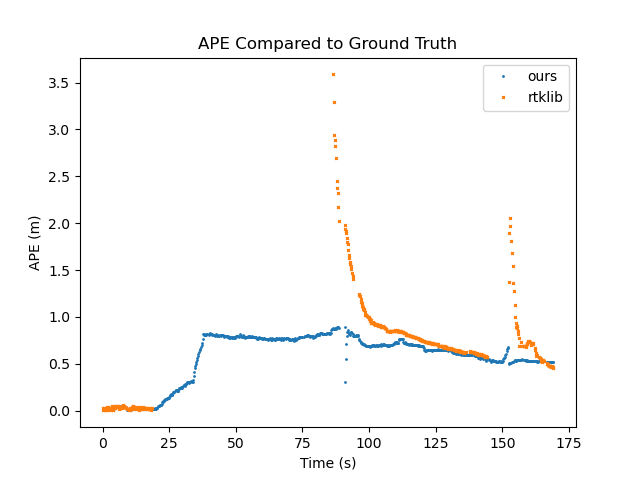}
      \caption{APE under open sky, with 10 satellites blocked, on average 4 satellites remain. RTKLIB fails some epochs.}
      \label{fig:uw-parking-lot}
 \end{subfigure} 
 \begin{subfigure}{0.33\linewidth}
     \centering
     \includegraphics[width=\linewidth,height=0.7\linewidth]{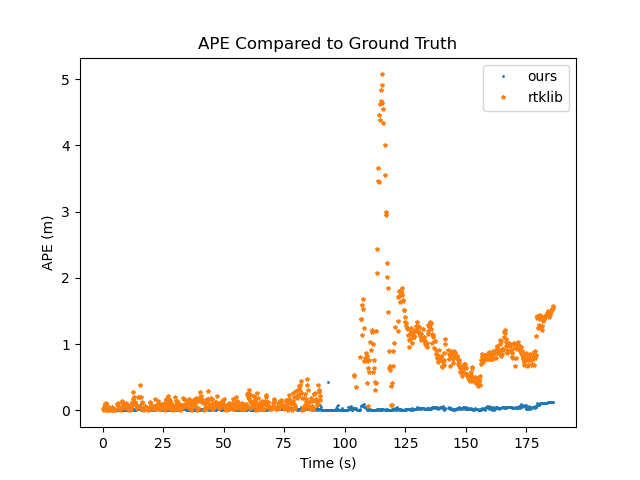}
     \caption{APE in lightly-blocked area, with 5 satellites blocked, on average 5 satellites remain visible.}
      \label{fig:redmond-survey}
 \end{subfigure}
  \begin{subfigure}{0.32\linewidth}
      \centering
      \includegraphics[width=0.8\linewidth,height=0.6\linewidth]{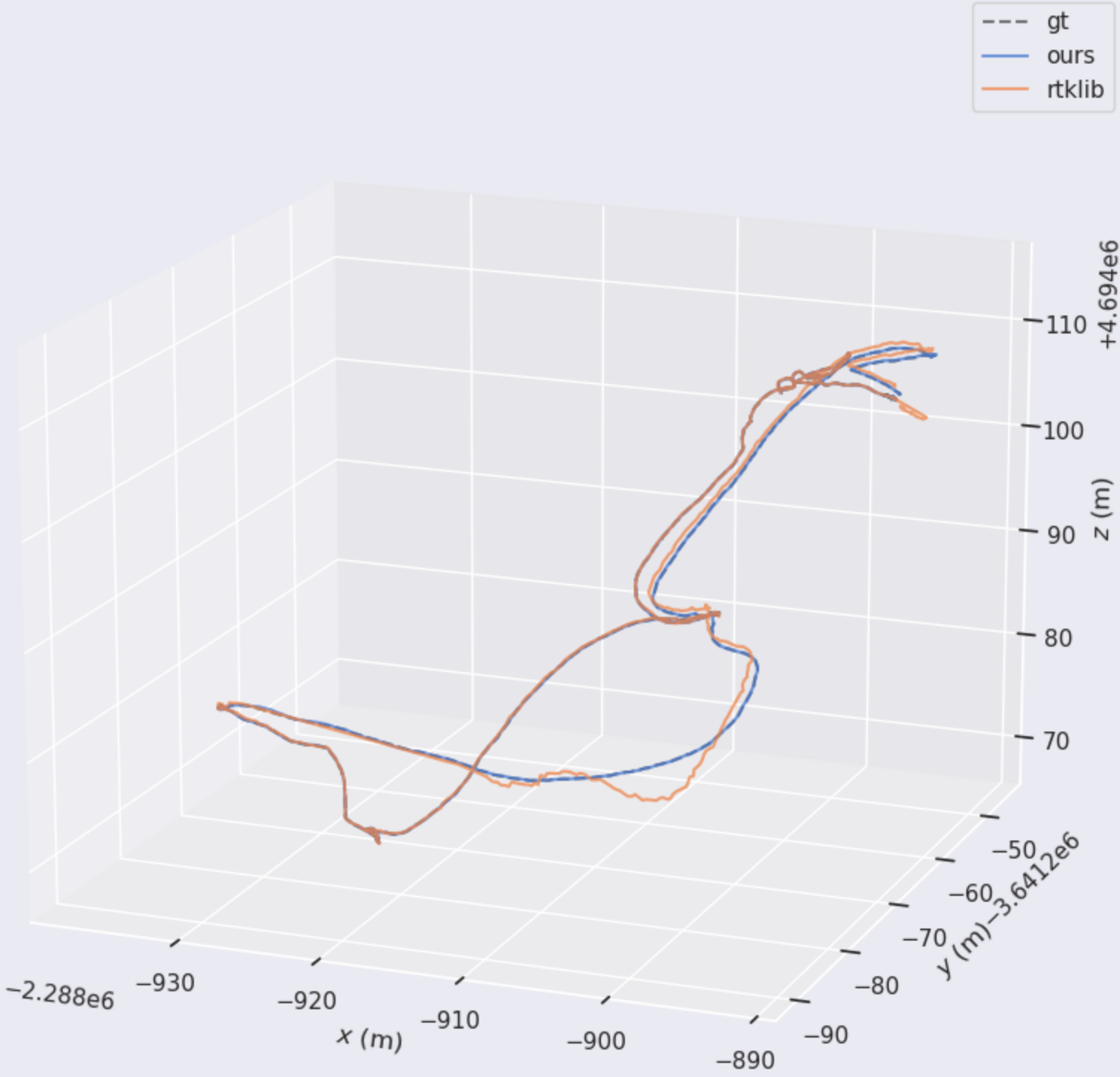}
       \caption{The comparison of the our trajectory, RTKLIB's trajectory, and the ground truth in ECEF frame in the lightly-blocked area}
       \label{fig:redmond-survey-traj}
  \end{subfigure}
 \caption{Performance comparison with ground truth for simulating signal blockage in urban canyons. Best viewed zoom in.} 
 \label{fig:simulation}
 \end{figure*}

\subsection{Analysis of Calibration for Alignment Accuracy}
\label{subsec:calib-on-alignment}
Table~\ref{tbl:alignment} shows alignment experiments with and without the proposed calibration for three trajectories, with lengths \SI{100}{\meter}, \SI{50}{\meter} and \SI{30}{\meter} respectively, under unobstructed sky visibility.
We can observe that the calibration significantly reduces the alignment error from decimeter-level to centimeter-level, up to \SI{12}{\cm} in trajectory 3, and thus largely boosts the alignment accuracy.
Another implication is that the alignment with the calibration has the same magnitude of error as carrier-phase positioning itself, and thus the accuracy of VIO and carrier-phase positioning are commensurate to be suitable for complementary sensor fusion.
\begin{table}[h]
\centering
\resizebox{\columnwidth}{!}{
\begin{tabular}{ c | c | c | c }
 & Traj 1 & Traj 2 & Traj 3 \\
 \hline
 Aligned Length (m) & 100 & 50 & 30 \\ 
 \hline
 RMSE without Calibration (m) & 0.11 & 0.11 & 0.16 \\  
 \hline
 RMSE with Calibration (m) & \bf 0.071 & \bf 0.028 & \bf 0.037 \\  
\end{tabular}}
\caption{Alignment errors with and without calibration}
\label{tbl:alignment}
\end{table}

Moreover, we show with a field experiment under open sky that GNSS pseudorange measurement is too noisy to provide VIO with a reliable centimeter-level ECEF-referenced registration. 
In Fig.~\ref{fig:qualitative}, the trajectory under good sky visibility is perceived by VIO, GNSS carrier-phase, and GNSS pseudorange.
We can observe that GNSS-PR is much noisier than the other two modalities, with a positioning error w.r.t. the other two modalities up to \SI{10}{\meter}.
This suggests that it is not suitable to fuse visual-inertial information with GNSS pseudorange because VIO cannot have centimeter-level registration to the ECEF frame by aligning to GNSS pseudorange positioning.
This conclusion is also supported by the ECEF registration error reported in Table II of GVINS~\cite{cao2022gvins}, where meter-level registration is observed even in an open-sky sports field.%

\begin{table}[htb]
\centering
\resizebox{0.85\columnwidth}{!}{
\begin{tabular}{ c | c  c  c  c  c  c}
 & Open Sky, 10 sats ex & Lightly-blocked, 5 sats ex\\
  &  RTKLIB : Ours & RTKLIB : Ours \\ 
  \hline
RMSE (m) $\downarrow$  &  0.68 : \bf 0.38  &  0.78 : \bf 0.033\\
Mean (m) $\downarrow$  &  0.55 : \bf 0.34  &  0.45 : \bf 0.025\\
Median (m) $\downarrow$  &  0.40 : \bf 0.32  &  0.26 : \bf 0.019\\
Std (m) $\downarrow$  &  0.41 : \bf 0.17  &  0.63 : \bf 0.023\\
Max (m) $\downarrow$  &  2.97 : \bf 0.88  &  4.90 : \bf 0.43\\
FSR (\%) $\uparrow$  & 10.8 : \bf 20.0 &  48.6 : \bf 92.9 
\end{tabular}}
\caption{Statistics of groundtruth comparison experiments. ``10 sats ex" means 10 satellites are manually excluded. Same applies to ``5 sates ex".}
\label{tbl:simulation}
\end{table}
%

\begin{figure*}[t]
\centering
\begin{subfigure}{0.245\linewidth}
    \centering
    \includegraphics[width=\linewidth,height=0.7\linewidth]{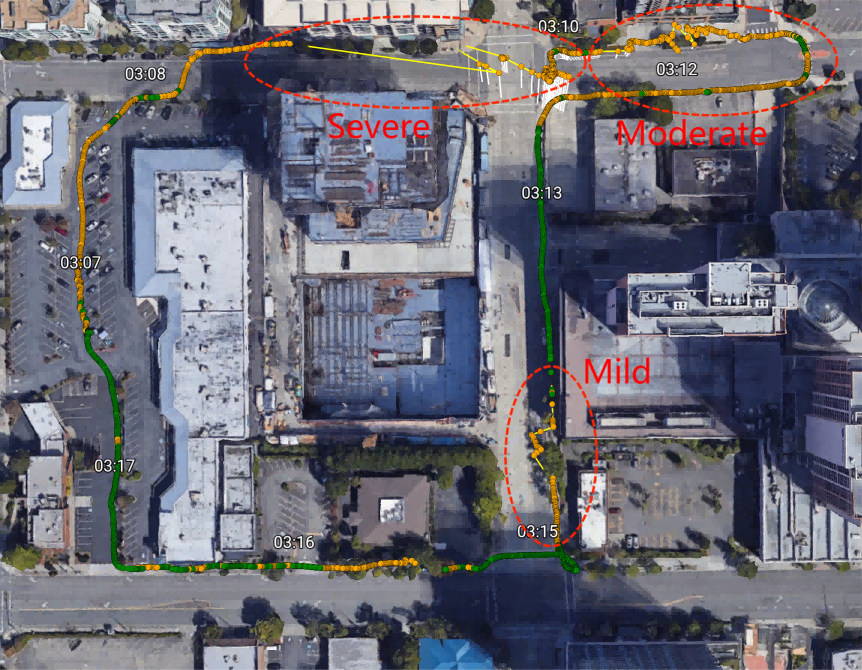}
    \caption{RTKLIB in UC1}
    \label{fig:downtown-original}
\end{subfigure}    
\begin{subfigure}{0.245\linewidth}
    \centering
    \includegraphics[width=\linewidth,height=0.7\linewidth]{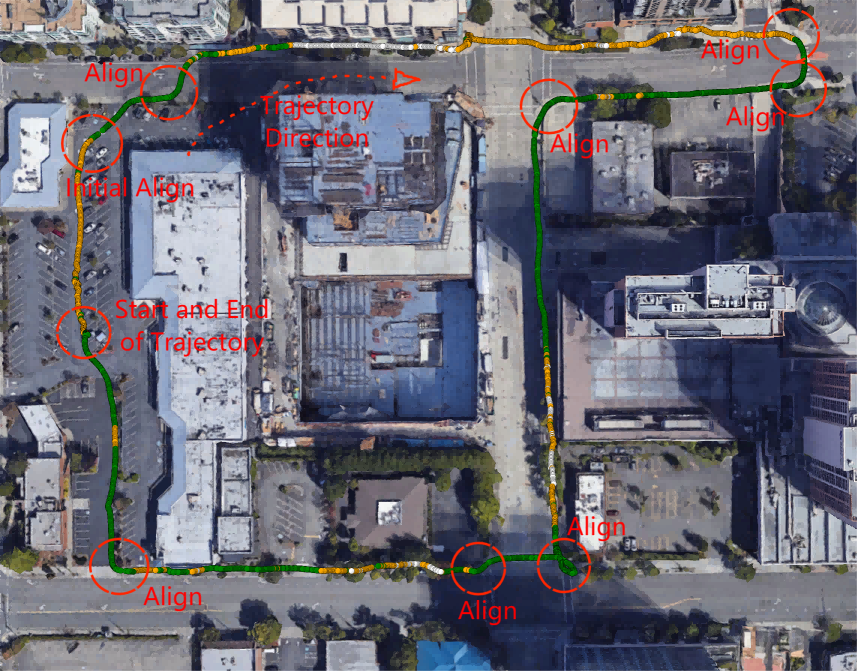}
    \caption{Our model in UC1}
    \label{fig:downtown-our}
\end{subfigure}
\begin{subfigure}{0.245\linewidth}
    \centering
    \includegraphics[width=\linewidth,height=0.7\linewidth]{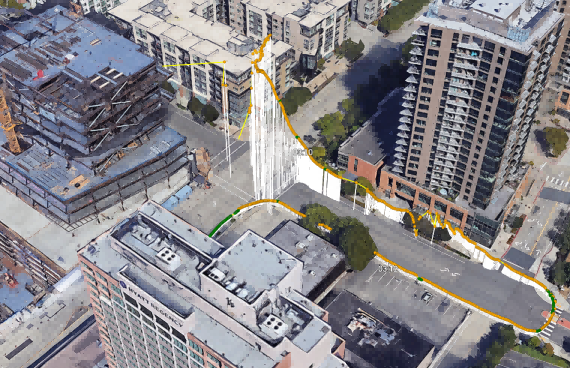}
    \caption{RTKLIB, UC1 (zoom-in view)}
    \label{fig:downtown-forward-zoom}
\end{subfigure}    
\begin{subfigure}{0.245\linewidth}
    \centering
    \includegraphics[width=\linewidth,height=0.7\linewidth]{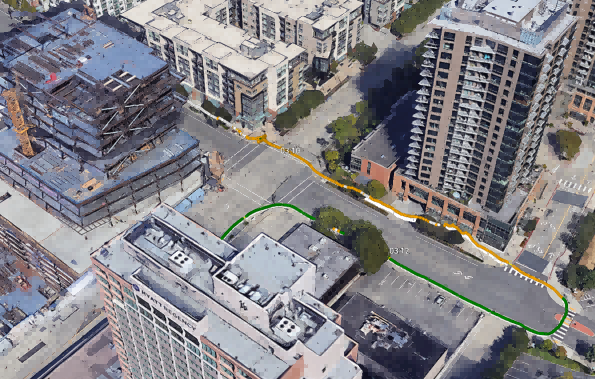}
    \caption{Ours, UC1's (zoom-in view)}
    \label{fig:downtown-causal-zoom}
\end{subfigure}
    
\caption{UC1 environment. Green: fixed solution. Yellow: float solution. White: aligned VIO when there is no valid float solution. Best viewed in color and zoom in.}
\label{fig:downtown}
\end{figure*}

\begin{figure*}[t]
\centering
\begin{subfigure}{0.245\linewidth}
    \centering
    \includegraphics[width=\linewidth,height=0.7\linewidth]{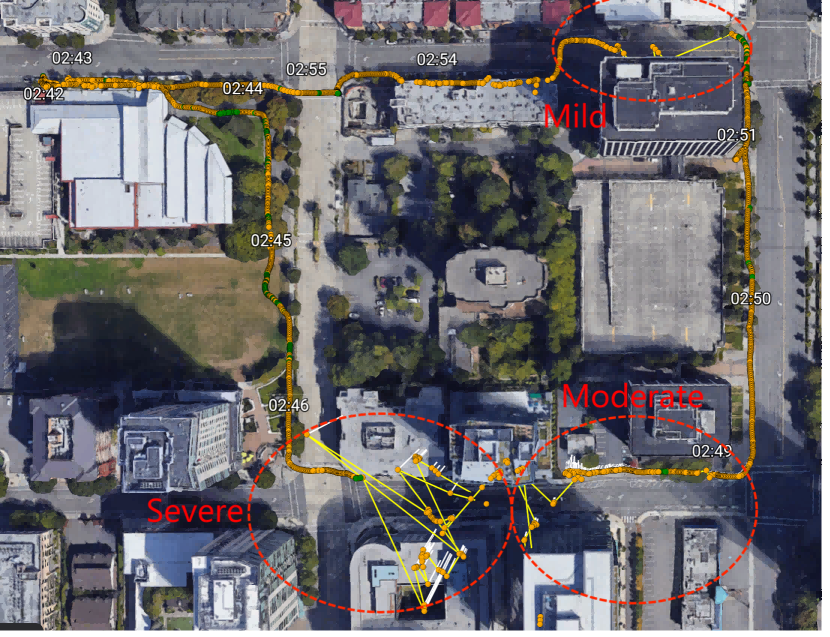}
    \caption{RTKLIB in UC2}
    \label{fig:library-original}
\end{subfigure} 
\begin{subfigure}{0.245\linewidth}
    \centering
    \includegraphics[width=\linewidth,height=0.7\linewidth]{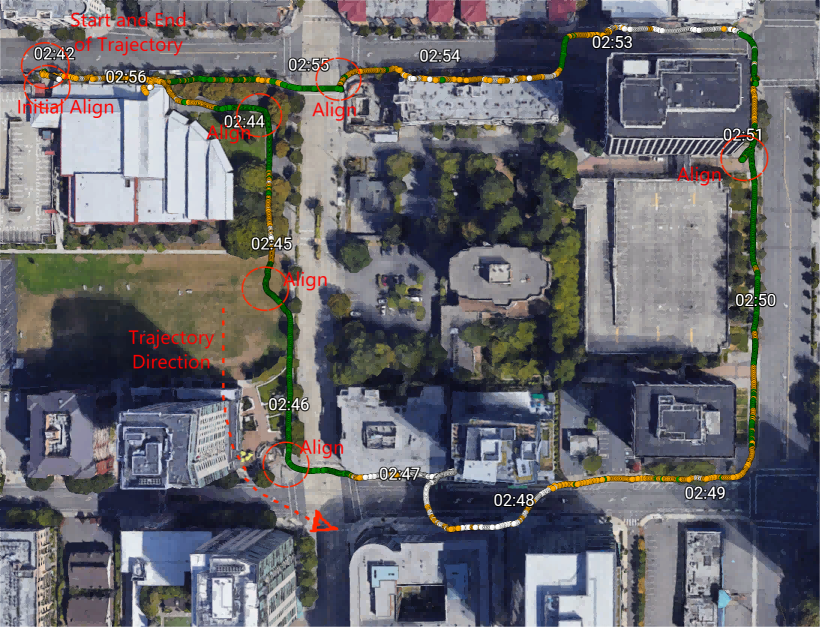}
    \caption{Our model in UC2}
    \label{fig:library-our}
\end{subfigure}
\begin{subfigure}{0.245\linewidth}
    \centering
    \includegraphics[width=\linewidth,height=0.7\linewidth]{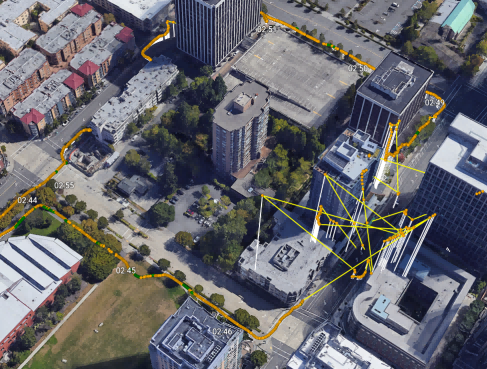}
    \caption{RTKLIB, UC2 (zoom-in view)}
    \label{fig:library-forward-zoom}
\end{subfigure}   
\begin{subfigure}{0.245\linewidth}
    \centering
    \includegraphics[width=\linewidth,height=0.7\linewidth]{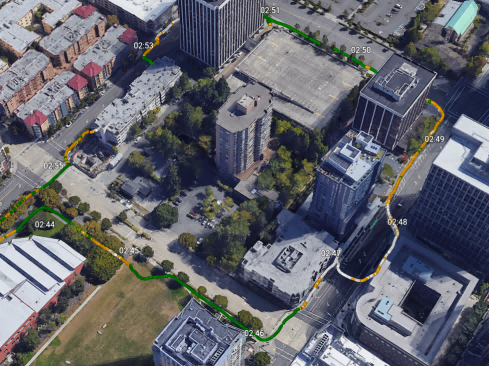}
    \caption{Ours, UC2 (zoom-in view)}
    \label{fig:library-causal-zoom}
\end{subfigure}
\caption{UC2 environment. Green: fixed solution. Yellow: float solution. White: aligned VIO when there is no valid float solution. Best viewed in color and zoom in.}
\label{fig:library}
\end{figure*}

\subsection{Groundtruth Comparison Experiment}
\label{subsec:simulation}
To characterize the performance of our method under severe sky blockage in urban scenarios, we need to be creative about how to get the groundtruth. It is also inappropriate to compare with methods that fuse low-accuracy pseudorange GNSS measurement into VIO~\cite{lee2020intermittent,qin2019general, cao2022gvins, boche2022visual}, which lack the ability to fuse carrier-phase GNSS measurement and instead directly use carrier-phase positioning as the groundtruth. To this end, we collected two datasets, one open-sky and one lightly-blocked sky by trees, and manually exclude satellites to simulate sky blockage, and then use RTKLIB to process full-constellation data as the groundtruth. 

We run our method and RTKLIB to process a subset of satellite measurements after manual satellite removal (10 satellites excluded for open-sky dataset and 5 excluded for lightly-blocked dataset).
The removed satellites contain the most frequently observed ones and moderately frequent ones during the observation period.
The absolute position error (APE) over time is shown in Fig.~\ref{fig:simulation} plotted by the evo toolkit~\cite{evo}.
%
The RMSE, mean APE, median APE, standard deviation of APE, max APE, and fixed solution rate (FSR, higher is better) are listed in Table~\ref{tbl:simulation}.
For both RTKLIB and our method, we use the same standard for the integer solution validation, which is the popular ratio test~\cite{verhagen2013ratio}, the ratio between the second best and the best residuals being 3.
We can observe that our method achieves lower position errors while having much higher fixed solution rate (FSR) than RTKLIB compared to the groundtruth.
\begin{figure}[H]
  \centering
  \includegraphics[width=0.7\linewidth]{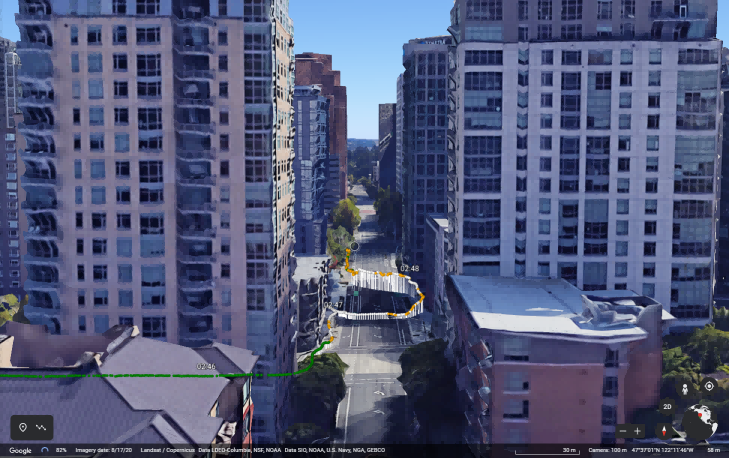}
   \caption{The urban canyons of the trajectory of Fig.~\ref{fig:library-our} visualized in Google Earth. The average building height in this canyon is 55m and the tallest building is 81m.}  
   \label{fig:uc-severe}
\end{figure}
\subsection{Real-World Urban Canyon Experiments}
\label{subsec:uc}
We conduct real-world experiments in two challenging urban canyon environments, ``UC1" and ``UC2" in Fig. \ref{fig:downtown} and Fig. \ref{fig:library}.
\footnote{Readers are also referred to https://youtu.be/PCDymbSTndQ for a video demonstration.}
A 3D street view of the urban canyon in our dataset is shown in Fig.~\ref{fig:uc-severe} to illustrate how challenging our problem is.
%
As is analyzed in Section~\ref{subsec:calib-on-alignment}, works that fuse GNSS pseudorange and VIO~\cite{lee2020intermittent,qin2019general, cao2022gvins, boche2022visual} are not suitable for comparison in our experiments, since their registration to ECEF have meter-level offset due to the noisiness of pseudorange positioning. 
Plus, they do not tackle carrier-phase positioning problem as we do.

%


%
In the trajectories, green means fixed solution, yellow means float solution, and white means ``fill-in" positions from aligned VIO when the validations in Alg.~\ref{alg:fusion} fails.
%
%
Visualizations of the results from RTKLIB and our model are shown in Fig.~\ref{fig:downtown} and Fig.~\ref{fig:library}.
We can observe that RTKLIB suffers from the urban canyon effects both horizontally (Fig.~\ref{fig:downtown-original},~\ref{fig:library-original}) and vertically (Fig.~\ref{fig:downtown-forward-zoom},~\ref{fig:library-forward-zoom}).
All the zig-zags and large jumps of the trajectory by the RTKLIB's EKF are eliminated by our model (Fig.~\ref{fig:downtown-our},~\ref{fig:library-our}).
Moreover, the segments formed by white spots connect smoothly with the GNSS positioning segments formed by green and yellow, which is a qualitative indicator that the our fusion method is coherent.
For quantitative evaluation, we pick three segments respectively from the UC1 and UC2 trajectories, where the urban canyon effects are severe, moderate, and mild, as marked in Fig.~\ref{fig:downtown-original} and Fig.~\ref{fig:library-original}.
%
%
%
In the selected segments, due to lack of ground truth, we use our method as a reference and evaluate RTKLIB's and the VIO's deviations from our method.\footnote{RTKLIB is commonly used as ground truth in SLAM research. In this sense, we are improving the ground truth used by many researches.}
The VIO trajectory is aligned with the selected segments by neighboring segments.
We show that although VIO has a drift rate of about $0.5\%$, it is much closer to our trajectory than RTKLIB's EKF, indicating that our system is much more robust than the original RTKLIB.
The RMSE, mean APE, median APE, and max APE are listed in the following two tables.
For simplicity, RTKLIB is denoted ``RTK" in the tables, and VIO denoted ``VIO".

\begin{center}
\noindent\resizebox{0.85\linewidth}{!} {%
\begin{tabular}{ c | c  c  c }
 UC1 Segments & Severe & Moderate  & Mild \\
  &  RTK : VIO & RTK : VIO & RTK : VIO \\
 RMSE (m) $\downarrow$ & 45.39 : \bf 1.61  & 9.59 : \bf 0.99 & 6.58 : \bf 1.11 \\ 
 \hline
 Mean (m) $\downarrow$ & 32.08 : \bf 1.29 & 4.79 : \bf 0.69 & 4.42 : \bf 1.02 \\  
 \hline
 Median (m) $\downarrow$ & 17.00 : \bf 0.96 & 1.32 : \bf 0.31 & 1.14 : \bf 0.94 \\
 \hline
 Max (m) $\downarrow$ & 82.06 : \bf 4.15 & 51.16 : \bf 3.30 & 14.94 : \bf 1.85
\end{tabular}
}
\end{center}

\begin{center}
\noindent\resizebox{0.85\linewidth}{!} {%
\begin{tabular}{ c | c  c  c }
 UC2 Segments & Severe & Moderate & Mild \\
  &  RTK : VIO & RTK : VIO & RTK : VIO \\
 RMSE (m) $\downarrow$ & 44.29 : \bf 2.33  & 23.21 : \bf 1.55 & 9.51 : \bf 0.79 \\ 
 \hline
 Mean (m) $\downarrow$ & 37.51 : \bf 1.88 & 14.69 : \bf 1.11 & 9.38 : \bf 0.63 \\  
 \hline
 Median (m) $\downarrow$ & 38.14 : \bf 1.02 & 5.57 : \bf 0.87 & 8.59 : \bf 0.59 \\
 \hline
 Max (m) $\downarrow$ & 112.44 : \bf 4.74 & 59.27 : \bf 4.10 & 12.40 : \bf 1.91
\end{tabular}
}
\end{center}

Finally, we compute the fixed solution rate of RTKLIB's EKF and our method in the following table.
We use the same ratio test standard for integer solution validation as Section~\ref{subsec:simulation}.
The results show that our method achieves significant improvement on fixed solution rate (higher is better), which means that our method has better robustness in urban canyons.

\begin{center}
\noindent\resizebox{0.8\linewidth}{!} {%
\begin{tabular}{ c | c  c  }
 Environments & UC1 & UC2 \\
  &  RTK : Ours & RTK : Ours \\
 \hline
 Fixed Solution Rate (\%) $\uparrow$ & 39.4 : \bf 50.8  & 8.1 : \bf 36.5 
\label{tbl:uc2}
\end{tabular}
}
\end{center}

%

%




%


\section{Conclusion}
This paper tackles the robustness issue of GNSS carrier-phase positioning in urban canyons with a loosely-coupled EKF to fuse GNSS carrier-phase and VIO.
We propose an optimization-based method to calibrate the extrinsics between IMU and GNSS antenna to accurately align VIO to the ECEF frame.
We use VIO as the prediction step for the EKF and incorporate its covariance estimate into the covariance propagation of EKF.
%
The performance of our method is quantitatively validated with simulation experiments and real-world experiments in challenging urban canyons.
Its robustness, positioning accuracy and fixed solution rate outperform the state-of-the-art GNSS carrier-phase positioning RTKLIB~\cite{takasu2009development} by a large margin in challenging urban canyons, filling the gap towards universally global positioning.
%
%
%

\bibliographystyle{IEEEtran.bst}
\bibliography{ref.bib}

\addtolength{\textheight}{-12cm}   









\end{document}